\documentclass[conference]{IEEEtran}
\IEEEoverridecommandlockouts
\usepackage{cite}
\usepackage{amsmath,amssymb,amsfonts}
\usepackage{algorithmic}
\usepackage{graphicx}
\usepackage{soul}
\usepackage{textcomp}
\usepackage{xcolor}
\usepackage{verbatim}
\usepackage[numbers, square]{natbib}
\def\BibTeX{{\rm B\kern-.05em{\sc i\kern-.025em b}\kern-.08em
    T\kern-.1667em\lower.7ex\hbox{E}\kern-.125emX}}
\begin{document}


\title{Position Paper: Integrating Explainability and Uncertainty Estimation in Medical AI}

\author{\IEEEauthorblockN{Xiuyi Fan}
\IEEEauthorblockA{
Lee Kong Chian School of Medicine,\\
College of Computing and Data Science,\\
Nanyang Technological University, Singapore}}



\maketitle

\begin{abstract}
Uncertainty is a fundamental challenge in medical practice, but current medical AI systems fail to explicitly quantify or communicate uncertainty in a way that aligns with clinical reasoning. Existing XAI works focus on interpreting model predictions but do not capture the confidence or reliability of these predictions. Conversely, uncertainty estimation (UE) techniques provide confidence measures but lack intuitive explanations. The disconnect between these two areas limits AI adoption in medicine. To address this gap, we propose \textbf{Explainable Uncertainty Estimation (XUE)} that integrates explainability with uncertainty quantification to enhance trust and usability in medical AI. We systematically map medical uncertainty to AI uncertainty concepts and identify key challenges in implementing XUE. We outline technical directions for advancing XUE, including multimodal uncertainty quantification, model-agnostic visualization techniques, and uncertainty-aware decision support systems. Lastly, we propose guiding principles to ensure effective XUE realisation. Our analysis highlights the need for AI systems that not only generate reliable predictions but also articulate confidence levels in a clinically meaningful way. This work contributes to the development of trustworthy medical AI by bridging explainability and uncertainty, paving the way for AI systems that are aligned with real-world clinical complexities.
\end{abstract}


\section{Introduction}
\label{sec:intro}

\begin{quote}
{\em ``Medicine is a science of uncertainty and an art of probability.''} -- William Osler    
\end{quote}

Medicine has always embraced the nuanced interplay of uncertainty and probability \cite{han2011varieties}. In modern healthcare, artificial intelligence (AI) holds immense promise for diagnosing diseases, personalizing treatments, and improving patient outcomes \cite{topol2019deep}. Yet, despite significant progress, the healthcare industry remains cautious about the reliability of AI-driven results \cite{kelly2019key}. This caution arises not merely from technical unfamiliarity, but from the core clinical reality that {\bf uncertainty} must be recognized, communicated, and managed effectively.

Explainable AI (XAI) has emerged as a key concept for illuminating how complex models arrive at predictions, especially in safety-critical domains like medicine \cite{tjoa2020survey}. However, while current XAI methods provide insights into \emph{why} a model makes a particular prediction by highlighting influential input features or reasoning pathways, they typically do not explain the uncertainty associated with these predictions. In other words, practitioners may understand the factors driving a model's decision but remain uninformed about the confidence or risk underlying that decision.

In AI research, Uncertainty Estimation (UE) techniques — ranging from Bayesian neural networks \cite{kendall2017uncertainties, blundell2015weight} to ensemble-based methods \cite{lakshminarayanan2017simple} — have steadily progressed. Yet their application in healthcare lags behind, partly because typical representations of uncertainty (often a single numeric confidence score) can feel opaque or disconnected from real-world decision flows \cite{bhatt2021uncertainty}. A surgeon, for instance, may be more interested in \emph{how} a system’s uncertain region was derived—does it stem from insufficient data on certain patient demographics, potential imaging artifacts, or intrinsic model variance?

This paper contends that next-generation medical AI must integrate comprehensive explainability \emph{and} robust uncertainty estimation to be truly trustworthy. We term this combined approach {\bf Explainable Uncertainty Estimation (XUE)}. By exposing not just the rationale behind a model’s decisions but also the underpinnings of its confidence, health professionals can better judge when to rely on automated insights and when to seek additional diagnostic opinions.

In this work, we make effort in the following directions.
\begin{itemize}
\item 
First, we present a mapping of medical uncertainties using state-of-the-art AI uncertainty estimation techniques, thereby defining the scope and need for XUE in medicine. 
\item 
Second, we identify the key challenges encountered in implementing XUE in medical AI, highlighting current research efforts aimed at addressing these obstacles. 
\item
Finally, we propose a set of design principles to inform future advancements in XUE, fostering dialogue between the Medical AI community and a broader interdisciplinary audience.
\end{itemize}

In the remainder of this paper, we offer a position on how the fields of explainability and uncertainty estimation should converge to realise XUE. We begin by surveying relevant literature on XAI and state-of-the-art uncertainty estimation in Section~\ref{sec:background}. Next, we study the relation between medical uncertainty and AI uncertainty in Section~\ref{sec:map}, drawing insights from both domains. Then, we discuss key challenges and research directions for XUE in medical AI, emphasizing its importance across diverse data modalities, clinical workflows, and human-centered evaluation frameworks in Section~\ref{sec:challenges}. Finally, we propose a set of guiding principles for developing XUE systems that enhance trust, interpretability, and decision support in high-stakes medical environments in Section~\ref{sec:principles}. 

\section{Related Work and Background}
\label{sec:background}

\subsection{XAI in Medical Applications}
\label{sec:xai}

In medical AI, ensuring model interpretability is paramount. XAI seeks to make the decision-making processes of AI systems transparent to users, particularly in high-stakes fields like healthcare \cite{tjoa2020survey}. We briefly review key XAI methodologies pertinent to medical applications, categorized into {\em Concept-based}, {\em Feature-based}, and {\em Example-based} explanations.

\subsubsection{Concept-Based Explanations}
Concept-based methods aim to align model interpretations with human-understandable concepts rather than low-level features~\cite{kim2018interpretability,ghorbani2019towards,bau2020understanding}. For instance, the \emph{Testing with Concept Activation Vectors} (TCAV) approach assesses a model's sensitivity to user-defined concepts by analysing directional derivatives in the activation space \cite{kim2018interpretability,de2024visual}. In medical imaging, this could involve evaluating a model's reliance on clinically relevant features such as ``tumor shape'' or ``lesion texture,'' thereby providing insights that are more intuitive for clinicians. Studies conducted in this direction include \cite{janik2021interpretability} for cardiac MRI segmentation.

\subsubsection{Feature-Based Explanations}

Feature-based explanations focus on identifying which input features most influence the model's predictions. Gradient-based methods, such as \emph{Layer-wise Relevance Propagation} (LRP) \cite{bach2015pixel}, \emph{Class Activation Mapping} (CAM) \cite{zhou2016learning}, and \emph{Gradient-weighted CAM} (Grad-CAM) \cite{selvaraju2017grad}, are widely used in medical imaging to produce saliency maps that localize critical regions in images. These methods have been applied to tasks like identifying pneumonia in chest X-rays \cite{rajpurkar2017chexnet} and detecting diabetic retinopathy in fundus photographs \cite{gargeya2017automated}.
Surrogate models approximate the behaviour of complex models using simpler, interpretable models. \emph{Local Interpretable Model-agnostic Explanations} (LIME) \cite{ribeiro2016should} builds local linear models around individual predictions to elucidate decision boundaries. Similarly, \emph{SHapley Additive exPlanations} (SHAP) \cite{lundberg2017unified} employs cooperative game theory to assign each feature an importance value for a particular prediction. These approaches have been utilized to interpret models predicting patient outcomes from electronic health records (EHRs) \cite{lundberg2020local}.

\subsubsection{Example-Based Explanations}

Example-based explanations provide insights by referencing specific instances from the data. Methods like \emph{Prototype Networks} \cite{li2018deep} identify representative examples (prototypes) and outliers (criticisms) from the dataset to explain model decisions. Prototype networks have been applied to various medical imaging tasks. For instance, \cite{wang2023novel} introduced a multimodal prototype network that integrates information from different imaging modalities to improve disease classification. \cite{choukali2024pseudo} applied prototype-based architectures to breast cancer classification. Additionally, prototype learning has been utilized for medical time series classification, such as electrocardiogram (ECG) signal analysis \cite{dakshit2024abstaining}. Counterfactual explanations involve generating minimally altered versions of an input that would change the model's prediction, helping in understanding decision boundaries and the robustness of the model. For example, \cite{wachter2017counterfactual} demonstrated how altering specific clinical features could change a diagnostic outcome, providing actionable insights for patient management. \cite{singla2023explaining} proposed a counterfactual approach using generative adversarial networks to modify chest X-ray images, illustrating how slight changes can flip a model's classification from ``diseased'' to ``healthy''. Similarly, \cite{tanyel2023beyond} explored counterfactual explanations in diagnosing paediatric brain tumours. \cite{tevsic2022can} investigated the impact of counterfactual explanations on users' causal beliefs, emphasizing the importance of careful implementation to avoid misconceptions.

\subsection{Uncertainty Estimation in AI Predictions}
Uncertainty estimation (UE) plays a pivotal role in deploying reliable, interpretable deep learning models \cite{gawlikowski2023survey}. By quantifying how confident a model is in its predictions, users and stakeholders can better gauge potential risks and decide when to seek further input or additional tests. Over the years, several methods have emerged to characterize different types of uncertainties, including \emph{aleatoric} (noise inherent to data) and \emph{epistemic} (model-related) uncertainty, as well as \emph{distributional} uncertainty (capturing shifts or anomalies in input distribution). Below, we briefly review existing approaches.

\subsubsection{Bayesian Neural Networks (BNNs)}
Bayesian Neural Networks treat model parameters as probability distributions rather than fixed values. This Bayesian formalism allows the simultaneous modeling of epistemic (model) and aleatoric (data) uncertainties~\cite{blundell2015weight, kendall2017uncertainties, maddox2019simple}. While BNNs can provide richer uncertainty estimates, they often require retraining from scratch and demand considerable computation, especially for large datasets. Post-hoc approaches such as BayesCap~\cite{upadhyay2022bayescap} mitigate these drawbacks by applying a Bayesian identity mapping to a \emph{pre-trained} deterministic model. Rather than learning the full model from the ground up, BayesCap transforms the original model’s outputs into well-calibrated uncertainties, offering a more computationally feasible route to Bayesian inference in deep neural architectures.

\subsubsection{Ensemble Methods}
Ensemble-based strategies assess prediction variance across multiple independently trained or differently parameterized models~\cite{lakshminarayanan2017simple, zou2023review}. The resulting variance provides a proxy for model uncertainty, often outperforming single-model approaches in reliability. However, ensembles can be computationally expensive to train, especially at scale, and their inference may be slower due to the need for multiple forward passes~\cite{durasov2021masksembles}. Monte Carlo (MC) Dropout is sometimes treated as a lightweight ensemble method: by retaining dropout at inference time, each forward pass effectively samples from a different sub-network~\cite{gal2016dropout}. The variance among these sampled predictions indicates the system’s uncertainty~\cite{leibig2017leveraging}. Although MC Dropout avoids training multiple distinct models, it still requires multiple forward passes during inference, increasing time costs~\cite{mi2022training}.

\subsubsection{Dirichlet-Based Methods}
Approaches grounded in the Dirichlet distribution model the class probabilities through a distribution-of-distributions perspective. Evidential Deep Learning (EDL) frames model outputs as ``evidence'' for a Dirichlet prior or posterior~\cite{sensoy2018evidential}, thereby decomposing total uncertainty into aleatoric and epistemic components. For instance, \cite{charpentier2020posterior} uses normalizing flows in the latent space to capture the Dirichlet parameters. Retraining models with Dirichlet assumptions can be computationally intensive~\cite{malinin2019reverse, malinin2020regression, nandy2020towards}. To address this, \cite{shen2023post} propose a post-hoc Dirichlet meta-model that leverages \emph{pre-trained} neural architectures to quantify uncertainty without requiring additional data. In parallel, \cite{yu2024discretization} improve EDL-based Auxiliary Uncertainty Estimators by introducing a Laplace distribution for heteroscedastic noise modeling and propose a Discretization-Induced Dirichlet Posterior (DIDO) for enhanced epistemic uncertainty estimation.

\subsubsection{Auxiliary Uncertainty Estimators (AuxUE)}
Auxiliary uncertainty estimation methods focus on obtaining uncertainty measurements without significantly altering the primary prediction task~\cite{hornauer2022gradient}. An early strategy uses \emph{entropy} derived from the softmax layer to represent model confidence: when uncertainty is high, 
entropy is high; conversely, low uncertainty corresponds to a more peaked probability distribution with low entropy \cite{pearce2021understanding}. \cite{wang2023uncertainty,korte2024confidence,hao2024exploring} leverage the reconstruction error—i.e., the distance to the training set—to quantify uncertainty. These strategies aim to minimize additional computational overhead while capturing reliable uncertainty indicators.

While XAI and UE have individually contributed to medical AI, their intersection remains underdeveloped. XAI methods provide insights into \emph{why} a model makes a prediction but do not indicate how reliable those explanations are. Conversely, UE techniques quantify uncertainty but fail to explain \emph{why} a given prediction is uncertain or where that uncertainty originates. This disconnect limits their clinical utility, as medical decision-making requires not only recognizing when an AI model is uncertain but also understanding the source and implications of that uncertainty. In the next section, we examine the nature of uncertainty in medicine and map it to AI uncertainties, defining the scope within which XUE operates.

\section{Mapping Uncertainties between Medicine \& AI}
\label{sec:map}

Uncertainty is an inherent and unavoidable aspect of medical practice, influencing clinical decision-making, patient care, and healthcare policy. Unlike other scientific disciplines, where uncertainty can often be reduced through additional data collection or improved models \cite{miskovic2019uncertainty}, medicine operates within a framework where some degree of uncertainty is inevitable \cite{kim2018understanding}. Despite the wealth of medical knowledge available today, clinicians continue to face uncertainty arising from three primary challenges. First, patient information is often incomplete, whether due to missing medical history, limited diagnostic tests, or variability in clinical presentations. Second, biological variability introduces an intrinsic level of unpredictability in disease progression and treatment responses. Third, medical conditions themselves are often complex, with multifactorial causes, co-morbidities, and atypical cases that defy straightforward classification. These factors mean that uncertainty cannot be eliminated but must be effectively managed \cite{helou2020uncertainty,simpkin2019communicating}.

Clinical practice has developed several strategies to cope with uncertainty, including multidisciplinary team discussions, iterative diagnostic testing, and shared decision-making with patients \cite{hamilton2016multidisciplinary, helou2020uncertainty}. Guidelines and decision-support frameworks help clinicians weigh risks and benefits under uncertain conditions. However, while these strategies offer practical solutions, they remain limited by the clinician's ability to interpret, communicate, and act on uncertainty. AI has the potential to augment this process, but only if it can quantify uncertainty in a way that aligns with real-world medical reasoning.

\subsection{A Taxonomy of Uncertainty in Medicine}

Han et al. \cite{han2011varieties} offer a systematic approach to understanding uncertainty in medicine through a three-dimensional taxonomy that classifies uncertainty according to its source, the aspect of care it affects, and who experiences it. 

The first dimension, \emph{source}, identifies where uncertainty originates. {\em Epistemic uncertainty} arises from incomplete knowledge, such as gaps in medical research, limited case studies, or ambiguous clinical guidelines. {\em Aleatory uncertainty}, in contrast, stems from inherent randomness in biological processes, such as variability in disease progression or unpredictable responses to treatment. A third category, {\em ambiguity}, emerges when medical information is conflicting or difficult to interpret, as seen in cases where multiple radiologists provide differing assessments of the same medical image.

The second dimension, \emph{issue}, examines what aspect of medical care is impacted by uncertainty. {\em Diagnostic uncertainty} occurs when symptoms and test results fail to definitively confirm a condition. {\em Prognostic uncertainty} affects predictions about how a disease will progress, particularly in cases with variable responses to treatment. {\em Treatment uncertainty} complicates decision-making regarding the best therapeutic approach, especially when data on effectiveness is limited. Uncertainty is also present at the systemic level, influencing healthcare policies, clinical guidelines, and medical workflows.

The third dimension, \emph{locus}, considers who experiences the uncertainty. {\em Clinicians} face uncertainty when interpreting ambiguous findings or making high-stakes decisions under time constraints. {\em Patients} experience uncertainty when trying to understand their diagnosis, potential outcomes, and treatment options. Healthcare institutions deal with uncertainty in managing resources, defining policies, and balancing risks across populations.

\subsection{Bridging Medical and AI Uncertainty}

The structured view of uncertainty in medicine maps naturally onto well-established AI uncertainty concepts. To ensure a coherent mapping, we explicitly connect each dimension of uncertainty—source, issue, and locus—to AI-driven modeling and explanation strategies.

\subsubsection{Source: Types of Uncertainty in AI and Medicine}

The three primary sources of uncertainty in medicine align with corresponding AI uncertainty categories:
\begin{itemize}
    \item \textbf{Aleatory uncertainty}, stemming from biological variability, corresponds to \emph{inherent noise} in AI models, such as uncertainty in diagnostic tests or imaging artifacts.
    \item \textbf{Epistemic uncertainty}, reflecting gaps in medical knowledge, aligns with \emph{model uncertainty} in AI, where the system is uncertain due to limited training data or unfamiliar cases.
    \item \textbf{Distributional uncertainty}, common in medical practice when clinicians encounter patient populations not well represented in studies, parallels \emph{out-of-distribution} uncertainty in AI, where an input deviates significantly from the model’s training data.
\end{itemize}
These uncertainties manifest differently depending on the type of data being analyzed. In electronic health records (EHRs), uncertainty often arises from missing values, inconsistent documentation, and heterogeneous patient populations \cite{holmes2021electronic}. Medical imaging introduces uncertainty due to variations in scanner equipment, acquisition protocols, and image quality, requiring AI systems to account for both aleatory noise and epistemic gaps in training data \cite{mccrindle2021radiology}. Time series data, such as ICU monitoring, poses additional challenges due to sensor drift, intermittent missing values, and evolving patient conditions, demanding uncertainty estimation methods that dynamically adapt over time \cite{singh2021missingness}.

\subsubsection{Issue: How Uncertainty Relates to AI Modeling}

In AI applications, \emph{issue} corresponds to the nature of the task the model is designed for. For instance, when AI systems are developed for \emph{diagnosis}, they must manage ``diagnostic uncertainty'', such as ambiguity in test results or imaging data. In contrast, AI models focused on \emph{prognosis} need to address ``prognostic uncertainty'', dealing with variability in disease progression and treatment responses. Recognizing these task-specific uncertainties allows AI models to be designed with domain-aware uncertainty quantification, ensuring they effectively support medical decision-making.

\subsubsection{Locus: Tailoring Uncertainty Communication in AI}

Here, \emph{locus} translates into the need for different explanation and communication strategies. Clinicians require AI models to convey uncertainty in a way that supports decision-making under time constraints, such as through confidence intervals or probability distributions. Patients, on the other hand, need AI-generated explanations that are interpretable and reassuring, helping them understand risk factors, prognostic uncertainty, and treatment options without unnecessary alarm. Ensuring that AI systems communicate uncertainty appropriately for each stakeholder group is essential for fostering trust and enabling effective decision-making in medical practice.

A key takeaway from this mapping is that AI models must not only quantify uncertainty but also explain it in a way that aligns with clinical reasoning and the needs of different audiences. In the next section, we explore key challenges and research directions for advancing this goal.

\section{Challenges and Research Directions}
\label{sec:challenges}

Five key challenges in delivering XUE are identified, ranging from technical implementation to result communication, and are summarized as follows.

\subsubsection{Challenge 1: Uncertainty Quantification Across Heterogeneous Data Modalities}
Given the different ways uncertainty manifests in clinical data, AI models must provide reliable estimates that appropriately reflect aleatoric, epistemic, and distributional uncertainty across diverse modalities for different applications. Electronic health records, often contain missing values and heterogeneous feature distributions, requiring models to estimate uncertainty at both the feature and patient levels. Medical imaging requires spatial uncertainty maps to highlight ambiguous regions. Time series data is complicated by sensor drift, intermittent missing values, and non-stationary dynamics, which demand uncertainty estimates that evolve over time. Ensuring consistency and interpretability across these modalities remains a fundamental challenge.

{\bf Solution Ideas:}  
To effectively quantify uncertainty across heterogeneous data modalities, methods must distinctly \ul{separate aleatoric, epistemic, and distributional uncertainties}. This distinction enhances interpretability and ensures reliable uncertainty estimation. In tabular data, Bayesian neural networks can be employed to model epistemic uncertainty, capturing the model’s confidence in its predictions, while aleatoric uncertainty can be assessed through variability in the data itself, such as measurement noise or missing values \cite{kendall2017uncertainties, abdar2021review}. In medical imaging, Monte Carlo dropout techniques during inference can estimate epistemic uncertainty, while aleatoric uncertainty can be quantified by modelling inherent noise in images, such as variations in image acquisition \cite{kendall2015bayesian,gast2018lightweight}. For time series data, recurrent neural networks with probabilistic layers can be considered \cite{rangapuram2018deep}. Across three modalities, reconstruction uncertainty estimation methods can be explored for measuring distributional uncertainty \cite{abdar2021review,korte2024confidence,wang2023uncertainty}. By explicitly modelling these different types of uncertainty and tailoring uncertainty quantification techniques to each modality, more robust uncertainty estimation can be achieved.

\subsubsection{Challenge 2: Communicating Uncertainty Effectively to Clinicians}
Even when AI models generate uncertainty estimates, their clinical utility remains limited because existing methods primarily present uncertainty only in the final result. This static representation fails to capture how uncertainty evolves throughout the model’s reasoning process, making it difficult for clinicians to assess the source of uncertainty. Existing XAI approaches for explaining predictions, such as saliency maps and feature importance scores, do not effectively convey confidence levels in a way that aligns with clinical reasoning. In radiology, pixel-wise uncertainty overlays can improve interpretability, while in EHR-based models, confidence intervals over risk predictions may offer a more in-depth understanding of uncertainty. To enhance clinical decision-making, uncertainty communication must go beyond presenting a single, final estimate and instead integrate into existing workflows in a way that dynamically supports clinician judgment rather than adding cognitive burden.

{\bf Solution Ideas:}  
Developing \ul{model-agnostic visualization tools} is crucial for bridging the gap between AI reliability and clinical intuition \cite{hullermeier2021aleatoric}. In \cite{bhatt2021uncertainty}, the authors discuss how uncertainty can be conveyed through visual, and linguistic representations, each suited to different audiences. Visual methods, such as confidence intervals, uncertainty overlays on images, and reliability diagrams, help interpret uncertainty in a structured format \cite{hullermeier2021aleatoric}. Linguistic representations, such as verbal probability statements (e.g., “highly uncertain” or “likely”), align with clinical communication but introduce ambiguity due to varied interpretations across users \cite{dhami2022communicating}. Finally, uncertainty-aware decision support systems that incorporate multi-level uncertainty representations tailored to specific clinical workflows can ensure that uncertainty is communicated in a way that is both actionable and clinically meaningful \cite{kompa2021second}.  

\subsubsection{Challenge 3: Evaluation of Explainable Uncertainty Estimation}
Despite advancements in uncertainty quantification, there is no established framework for evaluating the quality of explanations alongside uncertainty estimates. Metrics such as expected calibration error (ECE) and Brier scores assess numerical uncertainty accuracy but do not capture interpretability or clinical usability. The lack of standardized benchmarks for Explainable Uncertainty Estimation makes it difficult to compare methods and assess their effectiveness in real-world settings. A comprehensive and robust evaluation framework must incorporate both quantitative assessments and qualitative feedback from clinical users to ensure that uncertainty explanations are interpretable, trustworthy, and actionable.

{\bf Solution Ideas:} 
Evaluation frameworks must \ul{integrate quantitative and qualitative methods}. Metrics should assess the fidelity and stability of uncertainty explanations, ensuring they align with model behavior and remain consistent under input variations \cite{bhatt2021uncertainty, hullermeier2021aleatoric}. Qualitative assessments involving structured clinician feedback can further evaluate interpretability and decision-making utility \cite{kompa2021second}. Standardized benchmarks incorporating diverse medical datasets and uncertainty annotations are needed to enable systematic comparisons of XUE approaches. Finally, interdisciplinary collaboration among AI researchers, clinicians, and cognitive scientists is essential to develop rigorous, context-aware evaluation criteria \cite{johs2022explainable}. 

\subsubsection{Challenge 4: Integration of Domain Knowledge into Uncertainty Estimation}
Medicine is a knowledge-intensive field where clinical decision-making relies not only on patient data but also on clinical guidelines and expert reasoning. While purely data-driven approaches estimate uncertainty based on training data, they remain inherently limited by the quality, completeness, and representativeness of that data. As a result, uncertainty estimates may fail to capture rare conditions, evolving medical knowledge, or contextual factors. For instance, a model trained solely on structured EHR data may underestimate uncertainty in atypical presentations or emerging diseases, where expert interpretation is critical. To ensure uncertainty estimation aligns with clinical reasoning, uncertainty estimation methods must extend beyond training data by incorporating rich medical knowledge from clinical guidelines and textbooks to bridge the gap between statistical confidence and real-world medical judgment.

{\bf Solution Ideas:} 
One approach is to \ul{incorporate medical knowledge by leveraging large language models} to extract structured medical knowledge from guidelines and textbooks. This integration ensures that AI systems align with the latest evidence-based practices and estimate uncertainty based on up-to-date clinical knowledge \cite{hussain2020ai,sirocchi2024medical}. Another strategy is to develop hybrid modelling approaches that combine probabilistic frameworks with expert knowledge to capture both data-driven patterns and structured clinical reasoning. For instance, integrating Gaussian processes with deep learning models enables uncertainty quantification in treatment outcomes, facilitating personalized medicine and improving model robustness in rare or evolving conditions \cite{sun2022precision}. Finally, AI models should incorporate dynamic updating mechanisms that allow for continuous refinement of uncertainty estimation based on emerging medical knowledge. By designing adaptable frameworks that integrate new research findings and revised clinical guidelines over time, AI systems can maintain relevance and accuracy in rapidly evolving medical fields \cite{otokiti2023need}. 

\subsubsection{Challenge 5: Explaining Uncertainties in Generative and Large Language Models for Medical AI}
Generative AI and large language models (LLMs) are increasingly being used in medical applications, from automated clinical documentation to decision support and patient interactions \cite{yu2023leveraging}. However, LLMs introduce new challenges in uncertainty estimation due to their inherent stochasticity, reliance on pretrained knowledge, and susceptibility to hallucinations. Unlike traditional predictive models that provide structured outputs with confidence scores, generative models produce free-text responses or synthetic medical data, making it difficult to quantify and communicate uncertainty effectively. In medical AI, a lack of clear uncertainty indicators in generative outputs can lead to misinterpretation, overreliance on AI-generated insights, and potential patient harm. Additionally, generative models operate in open-ended domains where distributional shifts are more frequent, making conventional uncertainty estimation techniques insufficient. There is a critical need for methods that can quantify, explain, and communicate uncertainties in generative AI systems used in healthcare \cite{gao2025uncertainty}.

{\bf Solution Ideas:}
One approach is \ul{Confidence-Weighted Text Generation}, where AI models highlight uncertain phrases, provide confidence scores, and link generated outputs to verifiable sources. Techniques such as self-consistency sampling and retrieval-augmented generation (RAG) can help detect hallucinations and reduce uncertainty in free-text generation \cite{kadavath2022language,huang2023look,lin2023generating}. Interactive Uncertainty Exploration for Clinicians further enhances interpretability by allowing users to probe uncertainty, view alternative responses, or adjust model confidence thresholds. Interactive AI interfaces can display uncertainty overlays or offer explanation-driven re-generation options, preventing blind trust in AI outputs \cite{zhang2024rethinking,li2024trajvis}. Finally, Regulatory and Human-in-the-Loop Safeguards are critical for clinical deployment. AI systems should trigger human review for low-confidence outputs and align with regulatory standards for uncertainty-aware AI in healthcare. Establishing frameworks for AI oversight, including FDA-aligned safety mechanisms, ensures that uncertainty estimation is not just a technical feature but a clinical safety requirement \cite{mennella2024ethical}.

\section{Towards Robust Explainable Uncertainty Estimation: Key Principles}
\label{sec:principles}

Trust is the foundation of medical AI, and uncertainty estimation and explainable AI serve as complementary mechanisms to build and maintain that trust. Uncertainty estimation provides quantified measures of confidence, helping clinicians gauge the reliability of AI-generated insights. Explainability, in turn, ensures that these uncertainty estimates are transparent and interpretable, preventing blind reliance on AI-driven recommendations. Together, they synergize to create AI systems that are not only reliable but also understandable and actionable in clinical practice. Without explainability, uncertainty remains an opaque numerical artifact; without uncertainty estimation, explanations risk being misleadingly confident. By integrating these two paradigms, we can develop medical AI that supports, rather than replaces, clinical reasoning, ensuring that AI-driven decisions align with real-world complexities and clinician expectations.

To conclude this work, we propose the following principles for developing robust \emph{Explainable Uncertainty Estimation} systems in medical AI:

\begin{enumerate}
    \item \textbf{Clarity and Interpretability:}  
    Uncertainty explanations must be presented in a way that clinicians can intuitively understand. This requires translating model outputs into clinically relevant insights. Visual and textual explanations should be designed to minimize cognitive load, ensuring that uncertainty is not only quantified but also meaningfully contextualized.  

    \item \textbf{Traceability of Uncertainty:}  
    Effective uncertainty communication requires traceability, enabling clinicians to identify where uncertainty originates. Uncertainty scores should be linked to data quality issues, ensuring that uncertainty is not just reported but understood in context.

    \item \textbf{Actionability and Clinical Relevance:}  
    Uncertainty estimates should not merely describe confidence levels but also guide decision-making. Actionable AI systems should suggest additional tests, highlight cases requiring review, or provide alternative treatment pathways. AI-driven uncertainty must integrate with clinical workflows, ensuring that clinicians can respond meaningfully.

    \item \textbf{Human-Centered Design:}  
    The design and evaluation of XUE systems should prioritize clinician involvement. Iterative user feedback, participatory design, and usability testing are essential to ensure that explanations align with clinical reasoning processes. Uncertainty explanations should be tailored to domain-specific language and clinician expertise, ensuring that AI outputs are aligned with real-world decision-making needs.
\end{enumerate}

By adhering to these principles, we aim to bridge the gap between the inherent uncertainties in clinical practice and those modelled in AI systems. This approach helps ensure that AI is not merely a tool for prediction but a transparent, trustworthy partner in medical decision-making, capable of adapting to the complexities of real-world clinical environments.

\section*{Acknowledgement}
This research is supported by the Ministry of Education, Singapore
(RS15/23, LKCMedicine Start up Grant).

\bibliographystyle{ieeetr}
\bibliography{bibliography}

\end{document}